\documentclass[runningheads]{llncs}
\usepackage{graphicx}
\usepackage[table,dvipsnames]{xcolor} 
\usepackage{color}
\usepackage[]{todonotes}

\usepackage{tikz}
\usetikzlibrary{%
				automata,%
				arrows,%
				backgrounds,%
				calc,%
				chains,%
				decorations,%
				decorations.markings,%
				decorations.pathmorphing,%
				external,%
				fadings,%
				fit,%
				matrix,%
				positioning,%
				scopes,%
				shadows,%
				shapes.geometric,%
				shapes.misc,%
				shapes.multipart,%
				through,%
				mindmap, 
				backgrounds, 
				topaths}
		
\usetikzlibrary{arrows.meta}

\tikzset{
  pics/square/.default={1},
  pics/square/.style = {
    code = {
    \draw[pic actions, ultra thick] (0,0) rectangle (#1,#1);
    }
  }
}
	
\usepackage{tikz-3dplot}
\usepackage{units}

\usepackage{amsmath, amsfonts, amssymb}

\definecolor{KITred}{RGB}{160,30,40}

\definecolor{KITblue}  {RGB}{ 70,100,170} 
\definecolor{KITblue70}{RGB}{125,146,195} 
\definecolor{KITblue50}{RGB}{162,177,212} 
\definecolor{KITblue30}{RGB}{199,208,229} 
\definecolor{KITblue15}{RGB}{227,232,242} 
\definecolor{KITlilac}{RGB}{160,0,120}
\definecolor{KITpalegreen}{RGB}{130,190,60}
\colorlet{KITpalegreen70}{KITpalegreen!70}
\colorlet{KITpalegreen50}{KITpalegreen!50}
\colorlet{KITpalegreen30}{KITpalegreen!30}
\colorlet{KITpalegreen15}{KITpalegreen!15}
\definecolor{KITyellow}{RGB}{250,230,20}
\colorlet{KITyellow70}{KITyellow!70}
\colorlet{KITyellow50}{KITyellow!50}
\colorlet{KITyellow30}{KITyellow!30}
\colorlet{KITyellow15}{KITyellow!15}
\definecolor{Fraunhofergreen}{RGB}{23,156,125}
\definecolor{Fraunhoferblue}{RGB}{31,130,192} 
\definecolor{Fraunhofersteelblue}{RGB}{0,91,127}
\definecolor{Fraunhofersilvergrey}{RGB}{166,187,200}  
\definecolor{Fraunhoferorange}{RGB}{245,130,32}  
\definecolor{Fraunhofergraphit}{RGB}{28,63,82}  
\definecolor{Fraunhoferred}{RGB}{226,0,26} 
\definecolor{Fraunhofer_sand}{RGB}{211,199,174} 
\definecolor{Fraunhofer_weinrot}{RGB}{124,21,77} 
\definecolor{colormaplow}{RGB}{253,231,37} 
\definecolor{colormaphigh}{RGB}{68,13,84}

\usepackage[backref=page,pagebackref=true,breaklinks=true,colorlinks,bookmarks=false, linkcolor=Fraunhofergreen, citecolor=Fraunhofergreen, urlcolor=black]{hyperref}
\usepackage{subfig}
\usepackage{float}
\usepackage{pifont}%

\usepackage{multirow}
\usepackage[english]{babel}
\usepackage{breakcites}

\usepackage[]{todonotes}
\usepackage{amssymb}
\usepackage[detect-all,locale=DE]{siunitx} 

\newcommand{\cbox}[1]{\tikz[baseline=-0.5ex]\draw[#1, line width=3, ](0,0) -- (0.2, 0);}

\newcommand{\cmark}{\ding{51}}%
\newcommand{\xmark}{\ding{55}}%

\newcommand{\dmodel}{d_{\text{model}}}

\begin{document}
\title{MissFormer: (In-)attention-based handling of missing observations for trajectory \\ filtering and prediction}

\titlerunning{MissFormer: (In-)attention-based handling of missing observations}


\author{Stefan Becker \inst{1}\href{https://orcid.org/0000-0001-7367-2519}{\includegraphics[scale=0.04]{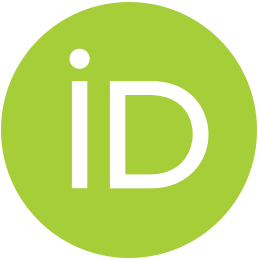}}\href{mailto:stefan.becker@iosb.fraunhofer.de}{\includegraphics[scale=0.04]{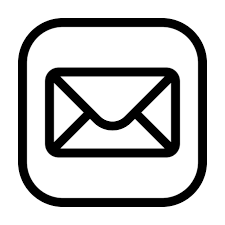}} \and
Ronny Hug \inst{1}\href{https://orcid.org/0000-0001-6104-710X}{\includegraphics[scale=0.04]{images/orcid-og-image.png}} \and 
Wolfgang Huebner \inst{1}\href{https://orcid.org/0000-0001-5634-6324}{\includegraphics[scale=0.04]{images/orcid-og-image.png}}   \and \\ Michael Arens \inst{1}\href{https://orcid.org/0000-0002-7857-0332}{\includegraphics[scale=0.04]{images/orcid-og-image.png}}
\and Brendan T. Morris\inst{2}\href{https://orcid.org/0000-0002-8592-8806}{\includegraphics[scale=0.04]{images/orcid-og-image.png}} } 
\authorrunning{S. Becker et al.}

\institute{Fraunhofer IOSB\thanks{Fraunhofer IOSB is a member of the Fraunhofer Center for Machine Learning.}, Ettlingen, Germany\\
\email{\{firstname.lastname\}@iosb.fraunhofer.de}\\
\url{www.iosb.fraunhofer.de} \and
University of Nevada, Las Vegas, USA\\
\email{brendan.morris@unlv.edu}}

\maketitle              
\begin{abstract}
In applications such as object tracking, time-series data inevitably carry missing observations. Following the success of deep learning-based models for various sequence learning tasks, these models increasingly replace classic approaches in object tracking applications for inferring the objects' motion states. While traditional tracking approaches can deal with missing observations, most of their deep counterparts are, by default, not suited for this.   

Towards this end, this paper introduces a \emph{transformer}-based approach for handling missing observations in variable input length trajectory data. 
The model is formed indirectly by successively increasing the complexity of the demanded inference tasks. Starting from reproducing noise-free trajectories, the model then learns to infer trajectories from noisy inputs. By providing missing tokens, binary-encoded missing events, the model learns to in-attend to missing data and infers a complete trajectory conditioned on the remaining inputs. In the case of a sequence of successive missing events, the model then acts as a pure prediction model. The abilities of the approach are demonstrated on synthetic data and real-world data reflecting prototypical object tracking scenarios.

\keywords{Transformer \and Trajectory Data \and Missing Input Data \and Filtering \and Trajectory Prediction \and Missing Observations}
\end{abstract}

\section{Introduction \& Related Work}
\label{sec:intro}
One crucial task for autonomous systems is estimating the agents' motion states based on observations. Following the success of deep learning-based models in various sequence processing tasks, like speech recognition \cite{chung2015recurrent,Graves_ICASSP_2013} and caption generation \cite{Donahue_CVPR_2015,Xu_MLR_2015}, these models are successfully utilized for trajectory prediction.

In trajectory prediction applications, deep learning-based approaches are increasingly replacing classic approaches due to their ability to capture better contextual cues from the static (e.g., obstacles; \emph{scene cues}) or dynamic environment (e.g., other objects in the scene; \emph{social cues})\cite{Rudenko_IJRR_2020}. Commonly used approaches for encoding object motions rely on \emph{recurrent neural networks} (RNNs) \cite{Alahi_CVPR_2016,Hug_RFMI_2017}, \emph{temporal convolution networks} (TCNs) \cite{Becker_ECCVW_2018,Nikhil_ECCVW_2018}, or \emph{transformers} \cite{Giuliari_ICPR_2020,Saleh_arXiv_2020}. The reader is referred to these surveys \cite{Rasouli_arXiv_2020,Rudenko_IJRR_2020,Kothari_arXiv_2020} for a comprehensive overview of current deep learning-based approaches for trajectory prediction. Since these models have the ability to consider \emph{social cues} and \emph{scene cues}, the focus of most research is how to incorporate these cues better. Although this research direction offers the strongest performance boost, problems such as missing observations are partly ignored or mainly addressed with data imputation and omitting the missing data \cite{Schafer_PM_2002}. To be more specific, this applies to trajectory prediction relying on observation extracted from an agent's trajectory as basic input (e.g., positions). Only this type of prediction problem is considered here and is referred to as \emph{trajectory cues}-based prediction in the remainder. For example, RNNs are designed to receive input data in every step and therefore are by default not suited to deal with missing inputs. In contrast, \emph{transformers} offer an alternative to the step-by-step processing in the form of the underlying attention mechanisms in combination with positional encoding.    
In general, data imputation means to substitute the missing values with methods like interpolation \cite{Kreindler_2006} or spline fitting \cite{DeBoor_2001} which results in a process where imputation and prediction models are separated \cite{Che_SREP_2018}. Thus, only suboptimal results are achieved since the model does not effectively explore the missing pattern. 
The simplest strategy for omitting is to remove samples in which a value is missing. While for RNNs this may work for training but cannot be applied during inference, omitting the missing value can be applied with \emph{transformers}. Giuliari et al. \cite{Giuliari_ICPR_2020} suggested omitting data with a \emph{transformer} model for trajectory prediction as an advantage compared to RNN-based models. They analyzed the effect of omitting the last observations of a fixed-length input sequence.   
Alternatively, and in particular for RNNs, the problem can be modeled with marked missing values. A missing value can be masked and explicitly excluded, or the model can be encouraged to learn that a specific value represents the missing observation (\emph{missing tokens})\cite{brownlee2017}. 
Most approaches are for healthcare applications \cite{Tresp_NIPS_1997} or in the field of speech recognition \cite{Parveen_NIPS_2001}. 
In the field of trajectory prediction, Becker et al. \cite{Becker_arXiv_2021} introduced an RNN-based full temporal filtering cycle for motion state estimation to better deal with missing observations. The Kalman filter-inspired model learns to weigh between its short-term predictions and observations enriched with missing tokens. In cases of missing inputs, the model entirely relies on predictions. Due to the recursive incorporation of new observations, deep Kalman models can be adapted similarly.\\ 
In this paper, we further explore the ability of \emph{transformer} networks to handle missing observations. Compared to the work of Giuliari et al. \cite{Giuliari_ICPR_2020}, we utilize a modified encoder-only \emph{transformer} model and provide missing tokens. Thus, the model is encouraged to learn specific placeholder values representing the missing observations. We analyze to what extent the combination of the underlying attention mechanisms with the positional encoding is able to then handle missing inputs along a variable length trajectory. Further, our model is not primarily designed as a prediction model, but in contrast, the model is formed indirectly by successively increasing the complexity of the demanded inference tasks. Starting from reproducing noise-free trajectories, the model then learns to infer trajectories from noisy inputs. The model outputs a full trajectory despite only being given partly observed trajectory data. Thus, for a sequence of successive missing events, the model then acts as a mere prediction model. The analysis of the model ability is performed under controlled conditions using synthetic data. For a comparison to other prediction models, the commonly used, publicly available \emph{BIWI} \cite{Pellegrini_ICCV_2009} and \emph{UCY} \cite{Lerner_CGF_2007} datasets are used.   

In the following, a brief formalization of the problem and a description of the proposed \emph{transformer} model are provided in section \ref{sec:model}. The achieved results are presented in section \ref{sec:evaluation}. Finally, a conclusion is given in section \ref{sec:conclusion}.

\section{MissFormer}
\label{sec:model}
The goal is to devise a model that can successfully infer the trajectory of a tracked agent conditioned on \emph{trajectory cues} (e.g., positions, headings, velocities) with missing observations. Trajectory prediction is formally stated as follows. Given an input sequence $\mathcal{X}$ of consecutively observed positions $\vec{x}^k=(p^k_{x},p^k_{y})$ (or other \emph{trajectory cues}) at time step $k$ along a trajectory, the task is to generate predictions for future positions $\{ \vec{x}^{k+1}, \vec{x}^{k+2},\ldots \}$
Here, we adapt the formal description as follows. Given a sequence of noisy, potentially missing observations $\tilde{\vec{x}}^{k}$, the task is to estimate the noise-free positions of the trajectory $\vec{x}^k$. So, $\tilde{\vec{x}}^{k}$ is a realization of $\vec{x}^k$ despite the fact the inputs of a \emph{transformer} are deterministic. In case the observations are noise-free, the task is to reproduce the trajectories. Although this might sound trivial, there exists no commonly accepted standards on encoding trajectory data in a deep learning model \cite{Hug_BTFM_2020}. When noise is present, the task is filtering. In addition to learning an adequate representation, the model needs to compensate input noise. 
If observations are missing at the end of the input sequence, the model acts as a prediction model and still infers the complete trajectory. Disregarding \emph{scene} and \emph{social cues}, trajectory prediction is here divided into different inference tasks with increasing complexity which the model can learn successively.\\

\textbf{Input/output:} For an agent, the \emph{transformer} network outputs the complete trajectory $\{ \vec{x}^{1}, \ldots \vec{x}^{k} \}$ up to time step $k$ conditioned on 
$\{ \tilde{\vec{x}}^{1}, \ldots \tilde{\vec{x}}^{k} \}$. To encourage the model to learn that a specific value represents missing, a binary-coded missing pattern is provided. The missing token is chosen as $(\vec{0}^{k},1^{k})$ for a missing observation and, respectively, $(\tilde{\vec{x}}^{k}, {0}^{k})$ for a default input, where $k\in \{1, k_{max}\}$. The adapted input is embedded onto a higher $\dmodel$-dimensional space by means of a linear mapping $\vec{e}^{k}= \text{EMB}(\tilde{\vec{x}}^{k}; \vec{\Theta}_{e})$. Accordingly, the output of the \emph{transformer} model is re-mapped to the $2$-dimensional coordinate system. Since \emph{transformers} contain no recurrence and no convolution, information about the position in the sequence must be injected. \emph{Positional encodings} are added to the input embeddings in accordance to the original \emph{transformer} \cite{Vaswani_NeurIPS_2017}. The \emph{positional encodings} have the same dimension $\dmodel$ as the embeddings, so that both can be summed up. Hence, the embedded input is time-stamped at time step $k$ by adding a \emph{positional encoding} vector $PE^{k}$. Following \cite{Vaswani_NeurIPS_2017}, sine and cosine functions of different frequencies are used to define $PE^{k}=\left\{ PE_{k,d} \right\}_{d=1}^{\dmodel}$ with
\begin{align}      
       PE_{k,d}=\left\{
        \begin{tabular}{lr}
        sin$\left(\frac{k}{10000^{d/\dmodel}}\right)$ & for $d$ even \\
        cos$\left(\frac{k}{10000^{d/\dmodel}}\right)$ & for $d$ odd 
        \end{tabular}
    \right. .
\end{align}

The time step $k$ corresponds to the position in the sequence and $d$ is the dimension. Each dimension of the positional encoding varies in time according to a sinusoid of different frequencies, from $2\pi$ to $10000 \cdot 2\pi$. That way, unique timestamps for sequences of up to 10000 elements are ensured.\\

\textbf{MissFormer:} Both the encoder and the decoder of a \emph{transformer} are composed of a stack of identical layers consisting of two sub-layers. Firstly, an attention module, and secondly, a feed-forward fully-connected module. Around each sub-layer, a residual connection followed by layer normalization is employed. Here, we only use the encoder and directly map the encoded state to an entire estimated trajectory instead of an auto-regressive generation with a decoder.    
Contrary to using a step-by-step processing of RNNs or convolution, \emph{transformers} rely entirely on self-attention to compute representations of its input and output. The attention function used by \emph{transformers} is the so-called scaled dot-product attention. The inputs consists of \emph{queries} and \emph{keys} of dimension $d_k$, and \emph{values} of dimension $d_v$ packed into matrices $Q$, $K$ and $V$. The attention layer is given by: 

\begin{equation}
   \mathrm{Attention}(Q, K, V) = \mathrm{softmax}(\frac{QK^\intercal}{\sqrt{d_k}})V
\end{equation}

\begin{figure*}[t!]
	\centering
	\resizebox{1.0\textwidth}{!}{
		\begin{tabular}{cc}			
			\includegraphics[]{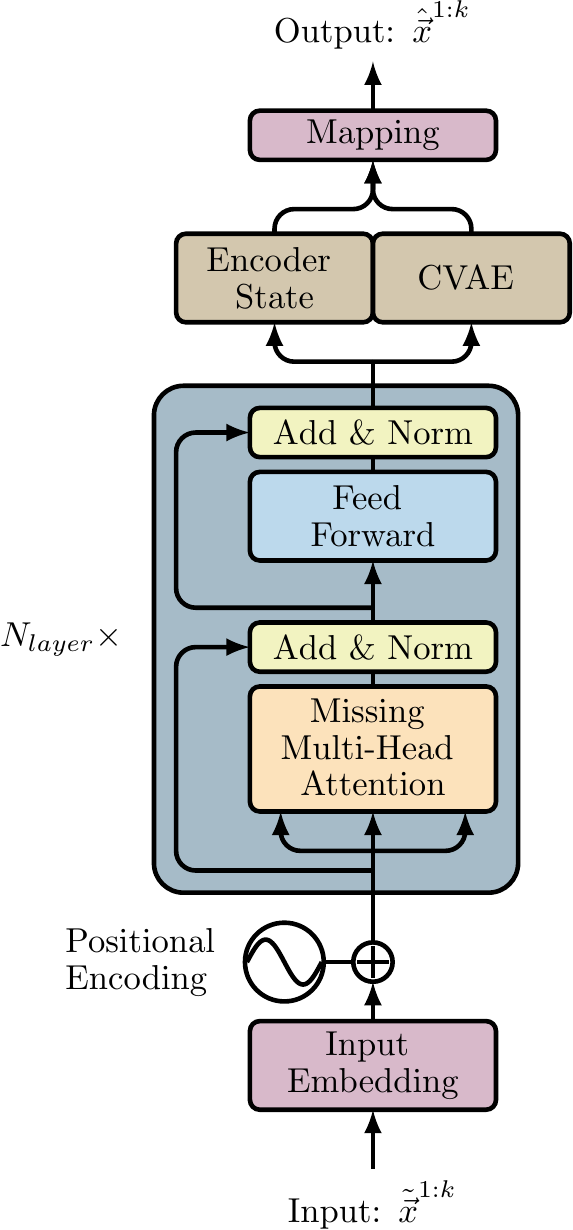} &
			\includegraphics[]{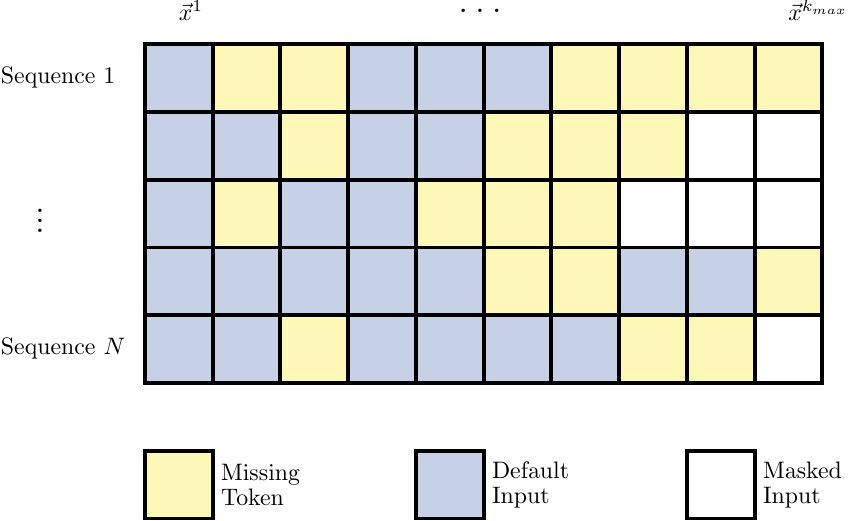} \\
		\end{tabular}
		}
	\caption{(Left) Visualization of the adapted \emph{transformer} network - MissFormer. (Right) Visualization of the adapted input data for the missing multi-head attention.}
	\label{fig:missfomer_mitoken}
\end{figure*}

So, the attention layer computes a dot product of the \emph{query} with all \emph{keys}, divided by $d_{k}$, and followed by a softmax function to obtain the weights on the values. Multi-head attention performs several attention functions in parallel, yielding to $d_{v}$-dimensional outputs. These values are concatenated before projected to the final value. The \emph{transformer} uses multi-head attention in different ways, whereas for an encoder-only architecture, solely the self-attention layer in the encoder is important. For more details, we refer to \cite{Vaswani_NeurIPS_2017}. In a self-attention layer of the encoder, all of the \emph{keys}, \emph{values} and \emph{queries} come from the same place, in this case the output of the previous layer in the encoder. On a high level, attention can be seen as routing of information. Thus, each position in the encoder can attend to all positions of the previous encoder layer or rather over all positions in the input sequence. The encoder creates a representation given the observation sequence resulting in the memory - the encoder state. 
In an encoder-decoder set-up, the encoder state is used to generate a \emph{key} and \emph{value} passed to the decoder. Since no new observations are provided to the decoder, we only used the encoder part. For every new observation, the extended input sequence is given to the model. As described, the model infers a sequence with a similar length to the input sequence. Our focus is on how well the described attention mechanism can route information to deal with missing observations and reconstruct complete, noise-free trajectories from the data. Further, the model is encouraged to still produce a meaningful representation and in-attend to useless placeholder values in the input sequence. The adapted \emph{transformer} model, referred to as MissFormer, together with modified input data with missing tokens, is visualized in Figure \ref{fig:missfomer_mitoken}.

On the left, the MissFormer with the missing self-attention sub-layer is shown. On the right, the \emph{missing tokens} are highlighted in yellow and default inputs are highlighted in blue. The Missformer is trained by minimizing the L$2$-loss in the form of the mean squared error between the ground truth trajectories and the estimated trajectories. Exemplary, the encoded state is combined with a conditional \emph{variational auto encoder}(VAE) \cite{Kingma_ICLR_2014} for producing multiple outputs and capturing the uncertainty of the estimation. Thus, the \emph{evidence-based lower bound} ELBO term is added as a second factor to the loss function (see for example \cite{Bishop_book_2006}). However, capturing the multi-modality of trajectory prediction is out of the scope of this paper. For example, this component can be replaced by \emph{flow}-based models  \cite{Dinh_ICLR_2014} or \emph{$\mathcal{N}$-curve} models \cite{Hug_AAAI_2020}.    

\section{Evalution \& Analysis}
\label{sec:evaluation}
This section consists of an evaluation of the proposed MissFormer. The evaluation concerns with verifying the approach's overall
viability in situations with missing observations in variable-length trajectory data. One part of the evaluation is done with synthetically generated data because, firstly, reference models can not handle mission observation by default. Secondly, current pedestrian trajectory data sets do not consider this aspect. For a comparison to other approaches, the publicly available \emph{BIWI} \cite{Pellegrini_ICCV_2009} and \emph{UCY} \cite{Lerner_CGF_2007} datasets are used according to the common practice of fully observed input data. 

\subsection{Synthetic Data}
\label{ssec:synthethic_data}
The synthetic data consists of diverse trajectories covering different types of prototypical object motion present in trajectory datasets \cite{Hug_IEEE_Access_2021}. The generated trajectories include the motion patterns of constant velocity, curvilinear motion, acceleration and deceleration motion.  
For generating synthetic trajectories of a basic object motion on a ground plane, random agents are sampled from a  uniform distribution of speeds ($\mathcal{U}(5.0\nicefrac{m}{s}, 10.0\nicefrac{m}{s} )$). The frame rate is set to $1{fps}$. The heading direction is sampled from $\mathcal{U}(\ang{0}, \ang{360})$ with a change of heading during a sampling period also sampled from uniform distribution of $\mathcal{U}(\ang{-20}, \ang{20})$. The de- and acceleration during a sampling period is sampled from $\mathcal{U}(-0.8\nicefrac{m}{s^2}, 1.5\nicefrac{m}{s^2})$. Missing events are drawn from a Bernoulli distribution $\mathcal{B}(\cdot,\cdot)$. The positional observation noise is assumed to follow a zero-mean Gaussian distribution $\mathcal{N}(0\nicefrac{m}{s}, (\cdot \nicefrac{m}{s})^2 )$. The evaluation set includes always $5000$ samples. The number of training samples is varied thought-out the experiments. The models have been implemented using \emph{Pytorch} \cite{Paszke_NEURIPS_2019}.  For training, an ADAM optimizer variant \cite{Kingma_ICLR_2015,Loshchilov_ICLR_2019} with a learning rate of $0.001$ is used. 

\begin{table*}[!h]
\resizebox{\textwidth}{!}{
	\begin{tabular}{| c c c c c c c c | c c |}
      \hline
      \hline
		 \multicolumn{10}{|c|}{Task: Encoding / Reconstruction} \\
			 \hline
			 input & output& obs. & pred. & {\#samples}&\#epochs & {noise} & {missing}  & {ADE}& $\sigma_{ADE}$  \\
			\hline			
			\rowcolor{KITblue!30}
      pos. & pos. & 8-20& \xmark & 1000 & 1000  & \xmark  & \xmark  & 0.067 & 0.013   \\
			\rowcolor{KITblue!30}
		 	off. & pos. & 8-20& \xmark & 1000 & 1000 &  \xmark  & \xmark  &  0.061 & 0.012   \\	
		 	\rowcolor{KITblue!30}
			pos. & pos. & 8-20& \xmark & 1000 & 1000 &  \xmark  & $\mathcal{B}(0.1,0.9)$  & 0.377 & 0.403     \\	
			\rowcolor{KITblue!30}
			off. & pos. & 8-20& \xmark & 1000 & 1000 &  \xmark  & $\mathcal{B}(0.1,0.9)$  & 0.175  & 0.154  \\			
			\rowcolor{KITblue!30}
			pos. & pos. & 8-20& \xmark & 3000 & 1000 &  \xmark  & $\mathcal{B}(0.1,0.9)$ & 0.138 & 0.074 \\
			\rowcolor{KITblue!30}
			off. & pos. & 8-20& \xmark & 3000 & 1000 &  \xmark  & $\mathcal{B}(0.1,0.9)$ & 0.155  & 0.079  \\
			\rowcolor{KITblue!30}
			pos. & pos. & 8-20& \xmark & 3000 & 3000 &  \xmark  & \xmark & 0.030  & 0.020   \\
			\rowcolor{KITblue!30}
			off. & pos. & 8-20& \xmark & 3000 & 3000 & \xmark  & \xmark &  0.039 & 0.013  \\
			\rowcolor{KITblue!30}
			pos. & pos. & 8-20& \xmark & 3000 & 3000 & \xmark  & $\mathcal{B}(0.1,0.9)$ & 0.087 & 0.060   \\
			\rowcolor{KITblue!30}
			off. & pos. & 8-20& \xmark & 3000 & 3000 &  \xmark  & $\mathcal{B}(0.1,0.9)$ &  0.095 & 0.065   \\				
			\rowcolor{KITblue!30}
			pos. & pos. & 8-20& \xmark & 4000 & 4000 &  \xmark  & \xmark & 0.028 & 0.015 \\
			\rowcolor{KITblue!30}
			pos. & pos. & 8-20& \xmark & 4000 & 4000 &  \xmark  &  $\mathcal{B}(0.1,0.9)$ & 0.081 & 0.015 \\
			\rowcolor{KITblue!30}
			off. & pos. & 8-20& \xmark & 4000 & 4000 &  \xmark  & \xmark & 0.031 & 0.014  \\
			\rowcolor{KITblue!30}
			off. & pos. & 8-20& \xmark & 4000 & 4000 &  \xmark  & $\mathcal{B}(0.1,0.9)$ & 0.084 & 0.014  \\
      \hline
			\hline
	   \multicolumn{10}{|c|}{Task: De-Noising / Filtering} \\
			 \hline
			 input & output& obs.  & pred.  & {\#samples}&\#epochs & {noise} & {missing}  & {ADE}& $\sigma_{ADE}$  \\
			\hline			
				\rowcolor{KITblue!30}
        pos. & pos. & 8-20& \xmark &   4000 & 4000 & $\mathcal{N}(0, 1^2)$ & \xmark  & 0.126 & 0.049   \\						
        \rowcolor{KITblue!30}
				pos. & pos. & 8-20& \xmark &  4000 & 4000 & $\mathcal{N}(0, 1^2)$ & $\mathcal{B}(0.1,0.9)$ & 0.165 & 0.071 \\	
				\rowcolor{KITblue!30}
				offs. & pos. & 8-20& \xmark &  4000 & 4000 & $\mathcal{N}(0, 1^2)$ & \xmark  & 0.148 & 0.055    \\	
				\rowcolor{KITblue!30}
				offs. & pos. & 8-20& \xmark &  4000 & 4000 & $\mathcal{N}(0, 1^2)$ & $\mathcal{B}(0.1,0.9)$ & 0.222 & 0.137  \\	
      \hline
			\hline
			\multicolumn{10}{|c|}{Task: Prediction} \\
			\hline
			input & output& obs.  & pred.  & {\#samples}&\#epochs & {noise} & {missing}  & {ADE}& $\sigma_{ADE}$  \\
			\hline
			\rowcolor{KITblue!30}
			pos. & pos.  & 8-14 & 6-12 &  4000 & 4000 & $\mathcal{N}(0, 1^2)$ & \xmark  & 0.809  & 0.514 \\
			\rowcolor{KITblue!30}
			pos. & pos. & 8-14 & 6-12 &  4000 & 4000 & $\mathcal{N}(0, 1^2)$ & $\mathcal{B}(0.1,0.9)$  &  0.920 & 0.422  \\
			\rowcolor{KITblue!30}
			offs. & pos. & 8-14 & 6-12 &  4000 & 4000 & $\mathcal{N}(0, 1^2)$ & \xmark  & 1.186  & 0.583 \\
			\rowcolor{KITblue!30}
			offs. & pos.& 8-14 & 6-12 &  4000 & 4000 & $\mathcal{N}(0, 1^2)$ & $\mathcal{B}(0.1,0.9)$  & 1.221 & 0.734 \\
			\hline
	\end{tabular} 
	}
	\caption{Results for a comparison between several trained MissFormer models for different inference tasks. The inference tasks are reconstruction, filtering and prediction. Reconstruction: In case the observations is noise-free, the task is to reproduce the trajectories. Filtering: Here positional observation noise is added and the model has to filter out this noise to generate noise-free trajectories. Prediction: Future object locations are inferred from noisy, observed trajectories.}
\label{tab:result-tf_ablation}
\end{table*}

In order to emphasize some statements, parts from the first experiments are summarized in Table~\ref{tab:result-tf_ablation}. For comparison of the different trained MissFormer models, the average displacement error (ADE) is calculated as the average L2 distance between the estimated positions and the ground truth positions. It should be noted that the model directly infers positions. Typically, only velocities or rather offsets are predicted and the last observation is used as a reference point. Since the amount of variation for offsets is lower compared to positions and the range in the data is more limited, less modeling effort and less data is required for model training (see from example \cite{Becker_ECCVW_2018}). Because the last observation is affected by noise or even missing, this practice is not applicable. Thus, here the outputs of the model are positions, and the inputs are varied between using positions or using offsets to infer positions by path integration.

The results show that the MissFormer is able to successfully in-attend to the missing tokens and successfully only uses the remaining inputs for conditioning. The difference between the model's estimate without missing observation and a missing probability $\mathcal{B}_{}(0.1,0.9)$ is very low. Of course, there is a drop in performance which can best be seen when looking at the reconstruction task. But the model can there basically learn a trivial solution of the identity of the input. Even when a trivial solution exists for using positions as inputs, the error does not drop to zero. Positional trajectory data can be seen as some sort of an increasing trend that cannot be fully captured by using non-linear activation functions. However, deep networks can achieve outputs greater than the bound of single activation functions, but they can saturate at minimum or maximum values, particularly for trending input data. For all inference tasks, but in particular for reconstruction, the model requires enough variation and training time to handle missing data. At first, the model simply reproduces placeholder values as outputs. By increasing the number of provided samples, the MissFormer starts to better route the information to compensate for outages. When the model has to additionally compensate for noise, the difference between missing and no-missing decreases. Here, the MissFormer has to generalize and filter out the noise. Thus, input identity mapping cannot be applied. When switching to a prediction task by replacing the last inputs with missing tokens, the difference between the models' estimates of fully observed and missing data decreases further.

In the experiments for all tasks, using positions as inputs works slightly better. The increased modeling effort is compensated by providing more variation during training. Since the error by using path integration for estimating the true position is propagated, the result is comprehensible. Further, in the context of a \emph{dynamical} system, only observing offsets is an unobservable system where it is impossible to identify the initial condition uniquely. Thus, the error in the first positional estimate cannot be compensated. However, without missing observations and low positional observation noise, choosing offsets over positions has shown superior results on public trajectory prediction benchmarks (\cite{Becker_ECCVW_2018}).    
   
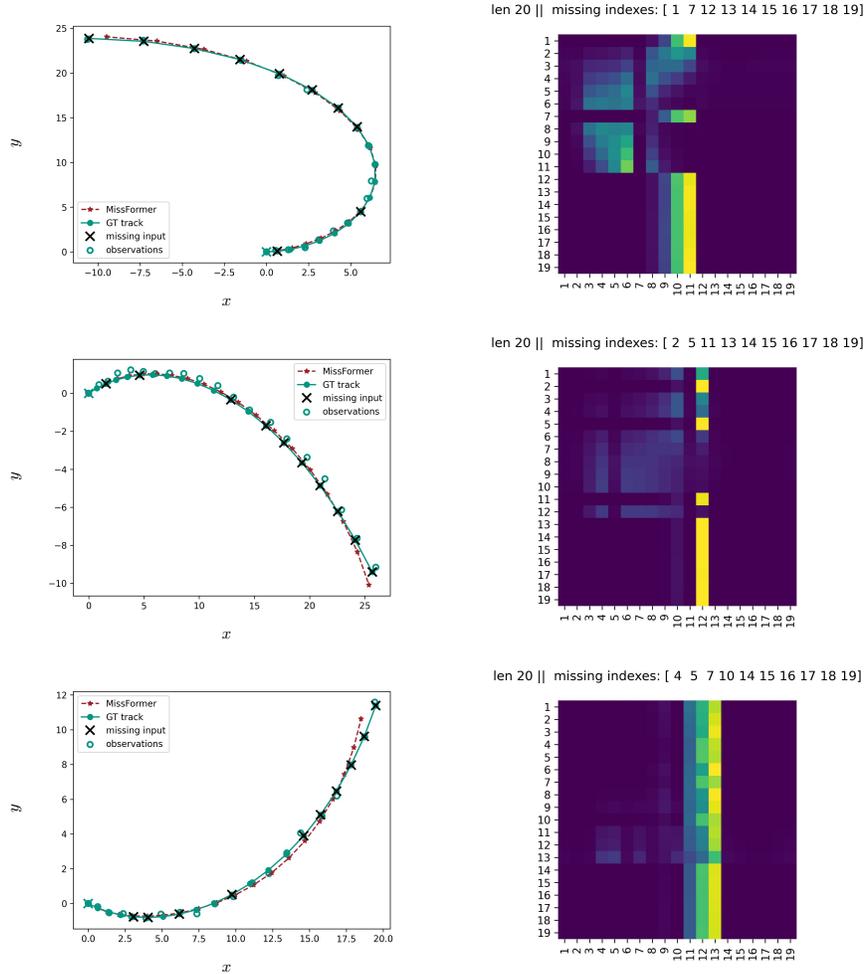
\begin{figure}[ht!]
\centering
	\resizebox{1.0\textwidth}{!}{
	\begin{tabular}{cc}				
				\begin{tikzpicture}
					\begin{scope}[font=\footnotesize]	
				\node (img1) at (0,0) {	\includegraphics[height=0.49\textwidth]{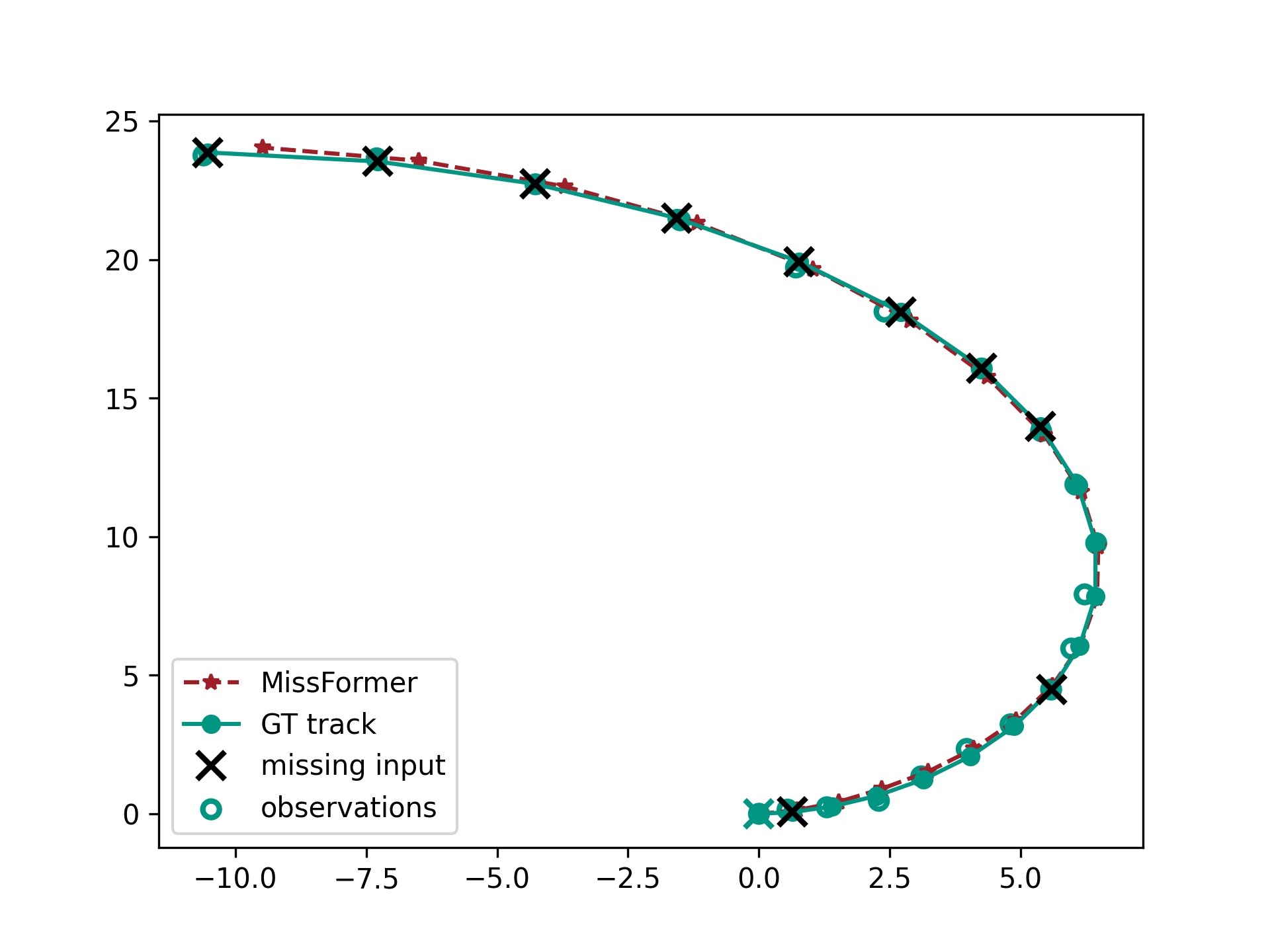}};
				\node [align=center] at (img1.south){$x$};  
		    \node [align=center,rotate=90] at (img1.west) {$ y$};
					\end{scope}									
				\end{tikzpicture} &
				\begin{tikzpicture}
				\begin{scope}[font=\footnotesize]
					\node (image2) at (0,0) {	\includegraphics[height=0.49\textwidth]{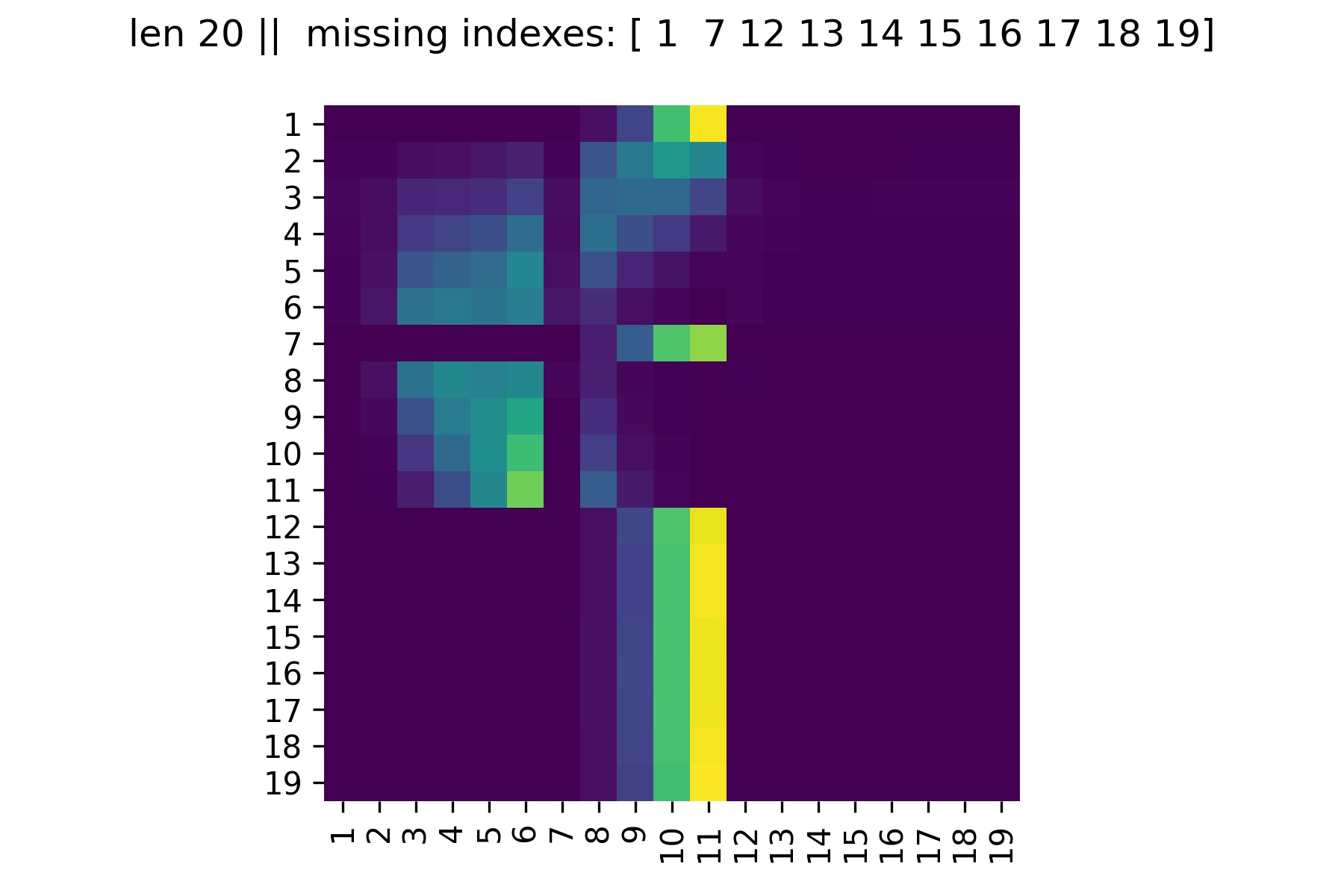}};			
					\end{scope}							
				\end{tikzpicture}	\\
					\begin{tikzpicture}
					\begin{scope}[font=\footnotesize]			
				\node (img1) at (0,0) {	\includegraphics[height=0.49\textwidth]{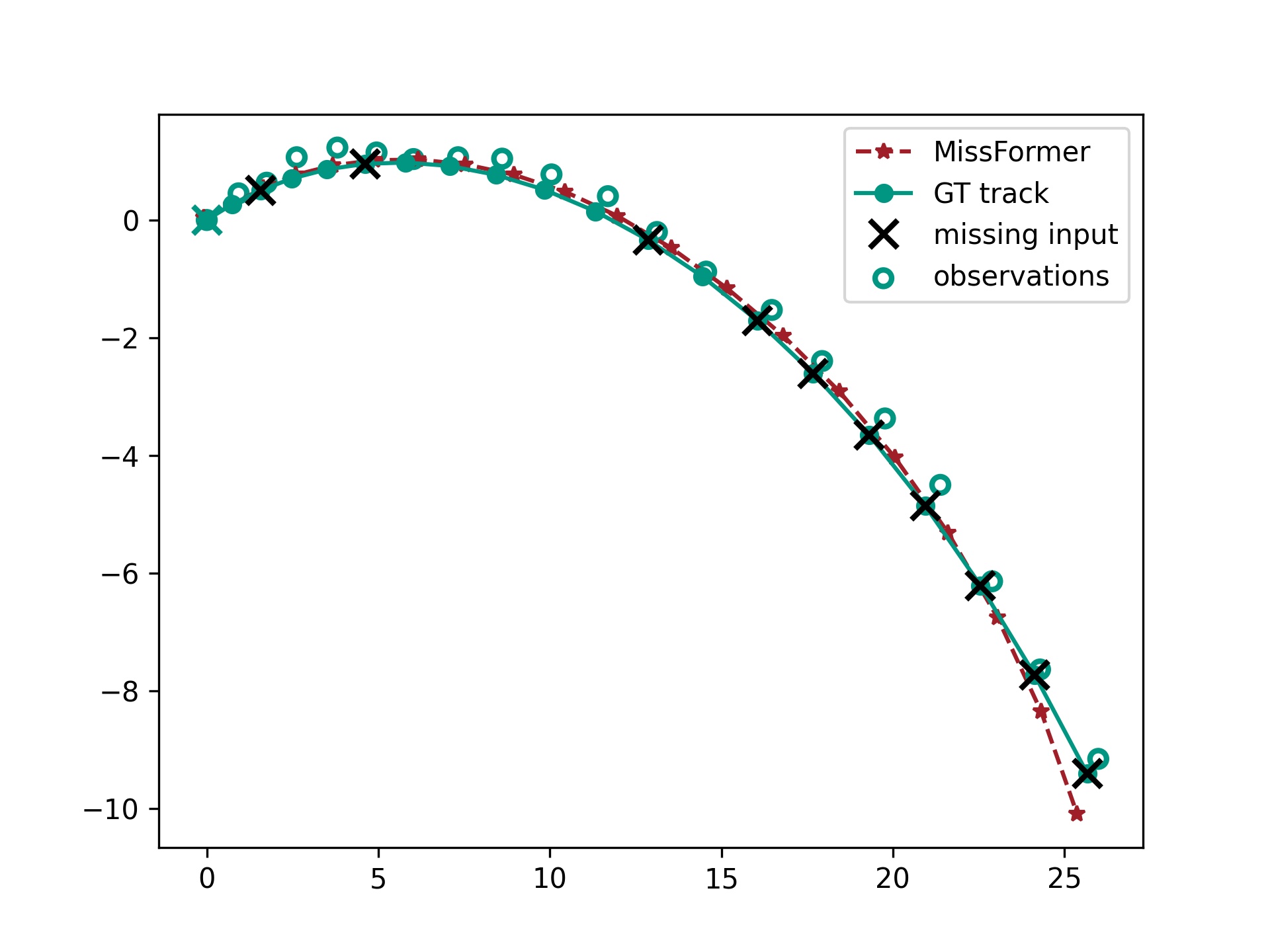}};
				\node [align=center] at (img1.south){$x$};  
		    \node [align=center,rotate=90] at (img1.west) {$ y$};	
					\end{scope}									
				\end{tikzpicture} &
				\begin{tikzpicture}
				\begin{scope}[font=\footnotesize]
					\node (image2) at (0,0) {	\includegraphics[height=0.49\textwidth]{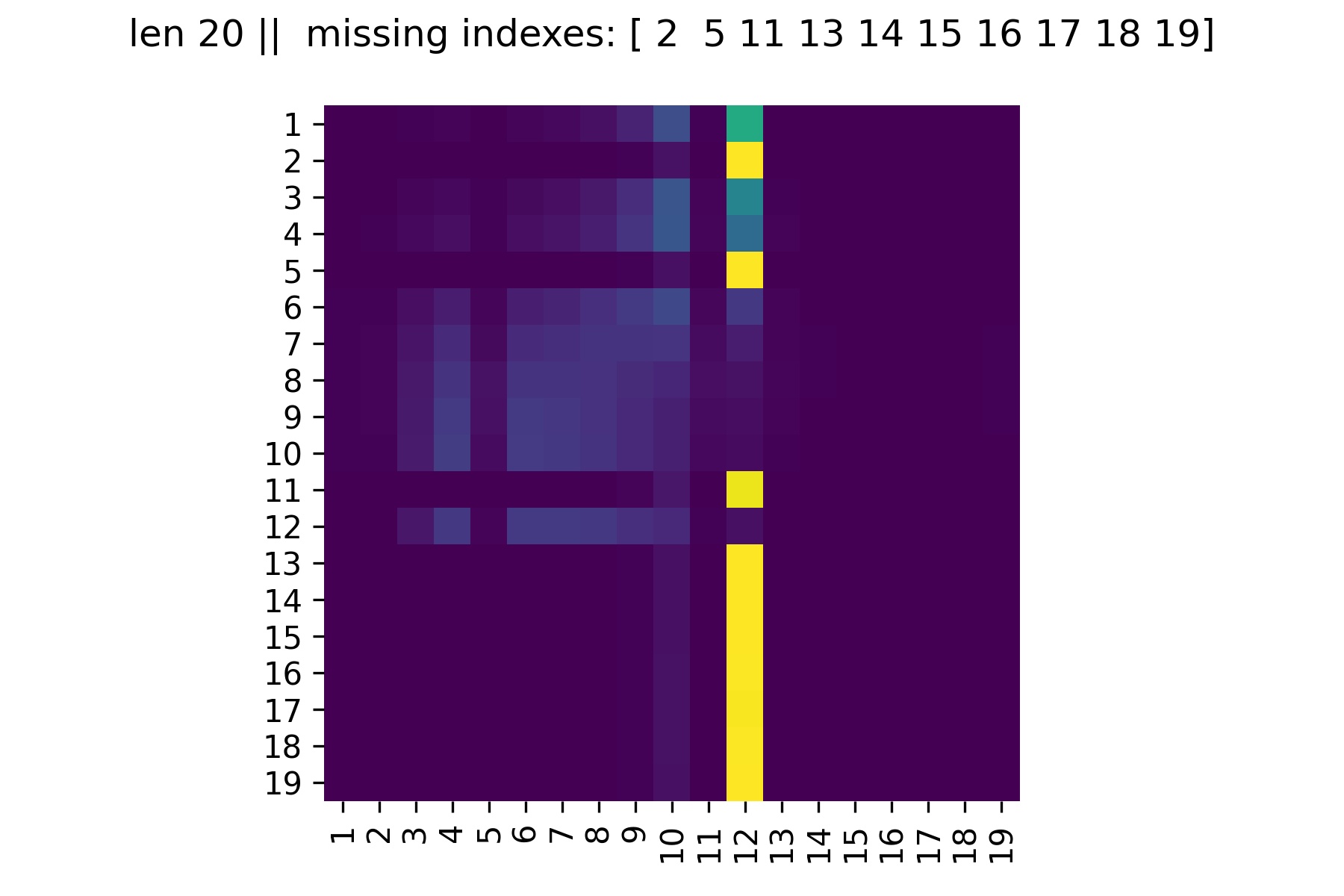}};		
					\end{scope}							
				\end{tikzpicture}\\
				\begin{tikzpicture}
				\begin{scope}[font=\footnotesize]			
				\node (img1) at (0,0) {	\includegraphics[height=0.49\textwidth]{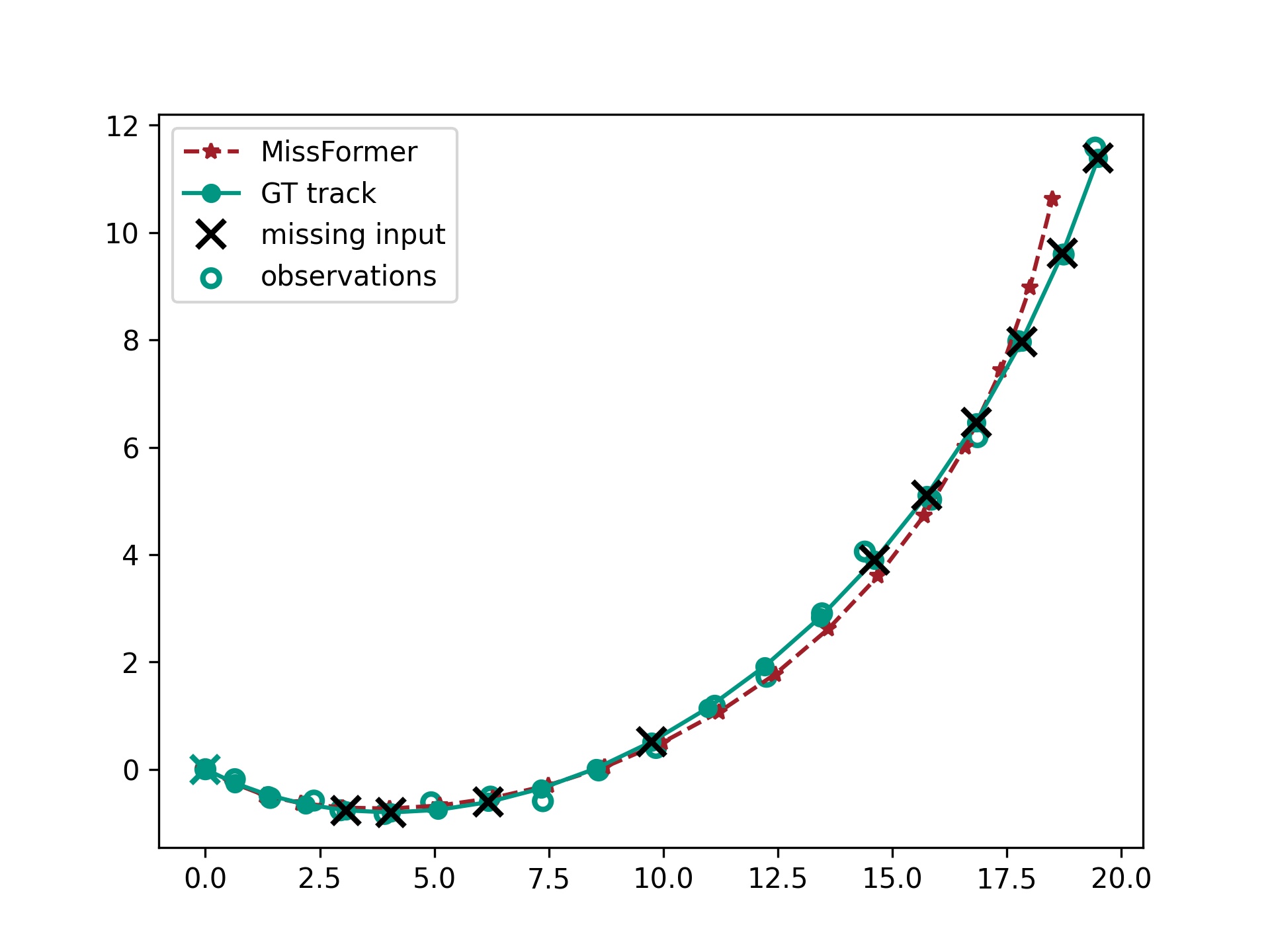}};
				\node [align=center] at (img1.south){$x$};  
		    \node [align=center,rotate=90] at (img1.west) {$ y$};
					\end{scope}									
				\end{tikzpicture} &
				\begin{tikzpicture}
				\begin{scope}[font=\footnotesize]
					\node (image2) at (0,0) {	\includegraphics[height=0.49\textwidth]{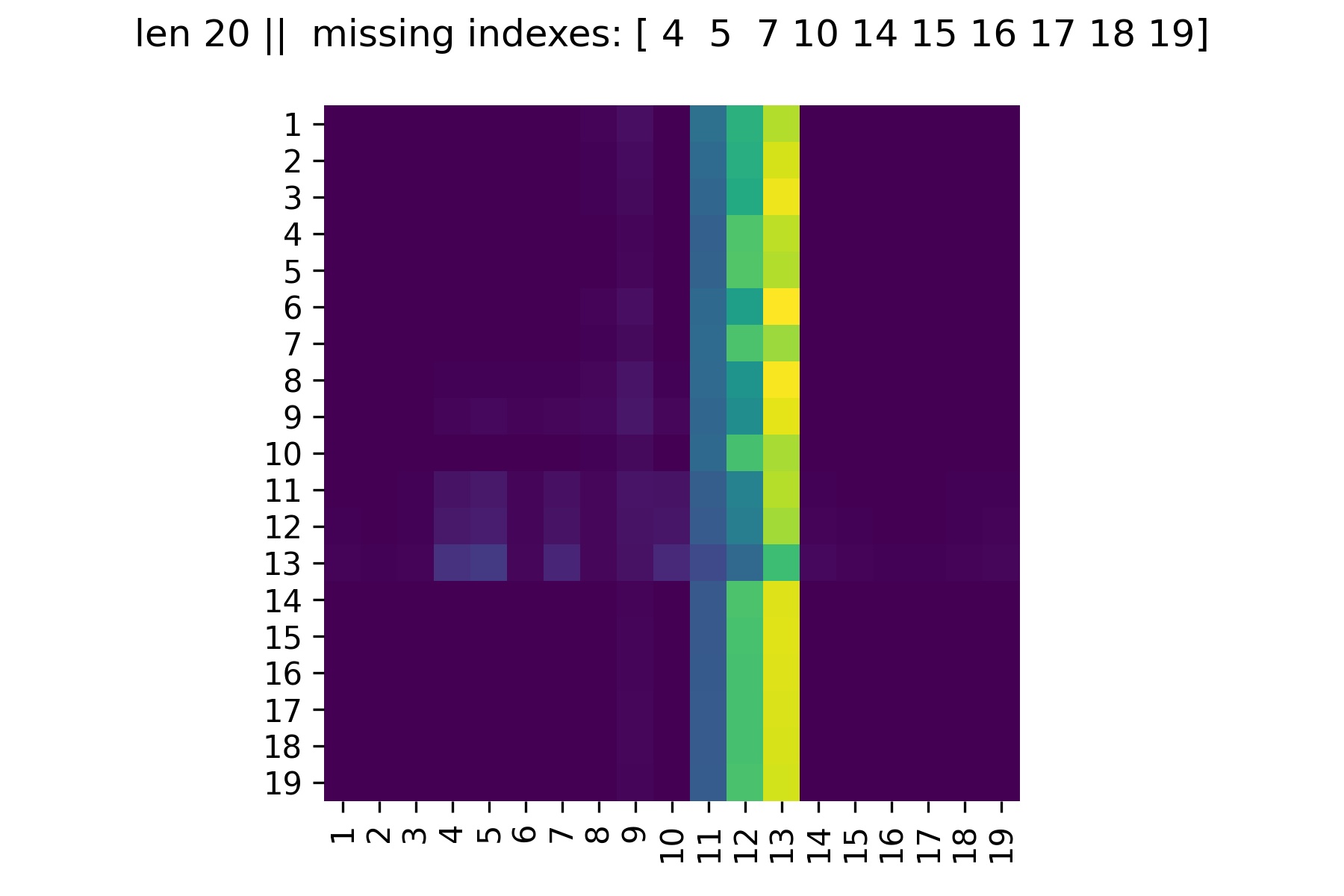}};	
					\end{scope}							
				\end{tikzpicture}				
  \end{tabular}}	
	\caption[]{(Left) Some exemplary estimates from the MissFormer with missing observations. The time steps where the input data is missing are marked with an cross. (Right) Corresponding attention filters. The attention filter values are color-coded ($0$\protect{\cbox{colormaphigh}}$\rightarrow$\protect{\cbox{colormaplow}} max.)}	
	\label{fig:example_attentions}
\end{figure}

For these experiments only one attention head ($N_{\text{head}}=1$) and one attention layer ($N_{\text{layer}}=1$) is used. Firstly, this allows a better understanding of the resulting attention because the attention filter directly shows what input information is used to encode a current trajectory. Secondly, compared to a \emph{nature language processing} (NLP) or a vision task, single trajectory processing requires no attention to several aspects of the input data (e.g., a second attention filter on context information in the background). Some exemplary estimates from the MissFormer with corresponding attention filters are depicted in Figure \ref{fig:example_attentions}. The values of the attention filter are color-coded ($0$\cbox{colormaphigh}$\rightarrow$\cbox{colormaplow}max.). Here, results for the prediction tasks are shown where the last inputs are purposely missing tokens. The time steps where the input data is missing are marked with a cross and missing input indexes are shown above the attention filters. The input length $k$ varies between $8$ and $14$ for a maximum length of $k_{max}=20$ with a missing probability of $\mathcal{B}_{}(0.1,0.9)$. 

The shown examples highlight several things. Firstly and most importantly, it can be seen that the models learned to in-attend to missing observation and to encode the 
trajectories based on the other inputs. Secondly, that the attention filters do not necessarily follow the typical look of high values along the diagonal as in an \emph{NLP} task. This can be explained by the fact that there are many possibilities for trajectory generation from the given inputs although solutions with predominantly high values along the diagonal can also result. Further, the examples show how the MissFormer filters out the noise and estimates relatively smooth outputs. And lastly they demonstrate, how prediction is done mainly relying on the last inputs, which corresponds to the common assumption that the last observations mainly influence motion. In summary, when providing the MissFormer with enough variation in terms of diversity and amount of training data, the model can handle missing observations and ignore placeholder values provided as missing tokens.     

\subsection{Real-World Data}
\label{ssec:real_world_data}
For real-world data and comparing the model to recent reference models, the publicly available \emph{BIWI} \cite{Pellegrini_ICCV_2009} and \emph{UCY} \cite{Lerner_CGF_2007} datasets are used. Combined, these datasets contain five sequences from an oblique view capturing scenes with pedestrians in a shopping street and on a university campus. As mentioned before, most reference models cannot handle mission observation and this problem is not considered in their corresponding evaluation. For the sake of completeness and in order to compare the approach to others, we follow the common trajectory prediction protocol. So, evaluation is done by leaving one-out cross-validation for the $5$ sequences. For conditioning, a fixed-length, fully observed trajectory of $8$ points ($3.2s$) is provided before predicting $12$ points ($4.8s$) into the future. The average displacement error (ADE) and the final displacement error (FDE) are used as error metrics. The ADE is defined as the average L2 distance between ground truth and the prediction over all predicted time steps and the FDE is defined as the L2 distance between the predicted final position and the actual final position. 

\begin{table*}[!h]
\resizebox{\textwidth}{!}{
  \begin{tabular}{|c|| c c c | c || c c c c c | c |}
      \hline
			\multicolumn{5}{|c||}{} &\multicolumn{6}{c|}{ADE/FDE in meters} \\
			 \hline
     	\multicolumn{1}{|c||}{Approach} & \multicolumn{3}{c|}{ cues } &\multicolumn{1}{c||}{model type}  & \multicolumn{5}{c|}{dataset  } &\multicolumn{1}{c|}{average } \\
			\multicolumn{1}{|c||}{} & {traj.}& social &  scene & \multicolumn{1}{c||}{}  &  \multicolumn{1}{c}{\emph{BIWI:ETH}}  & \emph{BIWI:Hotel} & \emph{UCY:Univ} & \emph{UCY:Zara1} & \emph{UCY:Zara2}  &  \\
			\hline
			\rowcolor{KITblue!30}
      Linear interpolation               &{\color{Fraunhofergreen}\cmark} & \xmark & \xmark & classic & 1.33/2.94 & 0.39/0.72    & 0.82/1.59 &0.62/1.21   &0.77/1.48  & 0.79/1.59    \\
			\rowcolor{KITblue!30}     
		 LSTM   & {\color{Fraunhofergreen}\cmark} & \xmark & \xmark  & RNN & 1.09/2.94 & 0.86/1.91  & 0.61/1.31  & 0.41/0.88  &0.52/1.11   & 0.70/1.52   \\			
			\rowcolor{KITblue!30}
			 GAN (Ind.)\cite{Gupta_CVPR_2018}  & {\color{Fraunhofergreen}\cmark} & \xmark & \xmark  & RNN & 1.13/2.21 & 1.01/2.18  & 0.60/1.28 & 0.42/0.91 &  0.52/1.11 & 0.74/1.54   \\	
			\rowcolor{KITblue!30}
     Social-LSTM  \cite{Alahi_CVPR_2016}  & {\color{Fraunhofergreen}\cmark} & {\color{Fraunhofergreen}\cmark} & \xmark  & RNN & 1.09/2.35 & 0.79/1.76  & 0.67/1.40 & 0.47/1.00 &0.56/1.17   & 0.72/1.54   \\
		 \rowcolor{KITblue!30}  
		 Social-Att. \cite{Vemula_ICRA_2018}  & {\color{Fraunhofergreen}\cmark} & {\color{Fraunhofergreen}\cmark} & \xmark  & RNN &  0.39/3.74 & 0.29/2.64 & 0.33/3.92 & 0.20/0.52 & 0.30/2.13 & 0.30/2.59    \\
		 \rowcolor{KITblue!30}  
		 Trajectron++\cite{Salzmann_ECCV_2020} & {\color{Fraunhofergreen}\cmark} & {\color{Fraunhofergreen}\cmark} & {\color{Fraunhofergreen}\cmark}  & RNN &  0.50/1.19 & 0.24/0.59 & 0.36/0.89 &0.29/0.72  & 0.27/0.67 &0.34/0.84      \\
		 \rowcolor{KITblue!30}  
		  TCN  \cite{Nikhil_ECCVW_2018}       & {\color{Fraunhofergreen}\cmark} & \xmark & \xmark & TCN & 1.04/2.07 & 0.59/1.17 &0.57/1.21 & 0.43/0.90 & 0.34/0.75 & 0.59/1.22   \\
			 \rowcolor{KITblue!30}  
			 TF   \cite{Giuliari_ICPR_2020}      & {\color{Fraunhofergreen}\cmark} & \xmark & \xmark & \emph{transformer} & 1.03/2.10 & 0.36/0.71 & 0.53/1.32 & 0.44/1.00  & 0.34/0.76  & 0.54/1.17 \\
			\rowcolor{gray!10}
       MissFormer (ours) &{\color{Fraunhofergreen}\cmark} & \xmark & \xmark  & \emph{transformer} & 0.99/1.94 & 0.36/0.89 &0.51/1.29 & 0.43/0.89 & 0.34/0.74 &  0.53/1.15   \\
      \hline
			\hline
	\end{tabular} 
	}
	\caption{Results for a comparison between the Missformer and a selection of recent prediction models following the single trajectory deterministic protocol. The prediction is done for 12 time steps into the future conditioned on 8 observations. Results are partly taken from \cite{Giuliari_ICPR_2020,Salzmann_ECCV_2020,Nikhil_ECCVW_2018}}
\label{tab:result-table-ETH-UCY}
\end{table*}

The results are summarized in Table \ref{tab:result-table-ETH-UCY}. In the comparison, a collection of recent approaches is considered where in terms of models relying solely on \emph{trajectory cues}    
at least one reference approach from the basic concepts of deep sequential trajectory processing and one classic approach is included (see column model type). The best performing models incorporate additional \emph{scene cues} (e.g., semantic segmentation), \emph{social cues} (e.g., interactions with other pedestrians) or both. When considering only \emph{trajectory cues}, the MissFormer achieves a better or similar performance. Without any outage in conditioning trajectory, the \emph{transformer} model of \cite{Giuliari_ICPR_2020} and our MissFormer model are very similar. Whereas Giuliari et al. utilize an encoder-decoder network with offsets as inputs and outputs, we use an encoder-only model with positional in- and outputs. To counter the lesser modeling effort of offset data, we pre-train the model on a diverse set of synthetically generated trajectories covering all types of prototypical pedestrian motion patterns. Therefore, the distribution and settings from section \ref{ssec:synthethic_data} for synthetic trajectory generation are adapted to match the underlying data better. For example, the frame rate is set to $2.5fps$ and random agents are sampled from a Gaussian distribution according to a preferred pedestrian walking speed \cite{Teknomo_phd_thesis_2002} ($\mathcal{N}(1,38\nicefrac{m}{s}, (0.37\nicefrac{m}{s})^2 )$). The model is pre-trained on a diverse set of $4000$ synthetic trajectories for $4000$ epochs. For the first half of training, the full trajectories are provided. Then, corresponding to the prediction length, the last inputs are replaced with missing tokens. Here, the number of heads and attention layer is set to $2$ ($N_{\text{head}}=2$, $N_{\text{layer}}=2$) and the model dimension is set to $\dmodel=256$.   
However, the achieved results for these datasets are very similar and the scope of this paper aims at further exploring the \emph{transformers}' ability to deal with missing observations.   
On the \emph{BIWI} \cite{Pellegrini_ICCV_2009} and \emph{UCY} \cite{Lerner_CGF_2007} datasets, there is no clearly best-performing individual \emph{trajectory cues}-based model. Overall, different models partly require different concepts for improving their performance or overcoming shortcomings. The presented results show that \emph{transformers} are a good choice for estimating trajectories and offer an built-in concept of dealing with missing inputs.  

\section{Conclusion}
\label{sec:conclusion}
In this paper, a \emph{transformer}-based approach for handling missing observations has been presented. The \emph{transformers}' built-in attention mechanisms in combination with positional encoding is analyzed in terms of exploring the remaining inputs for inference with outages. By providing encoded missing information (\emph{missing tokens}), the model is encouraged to learn that specific values represent missing. The presented results show that the model can in-attend to the placeholder values and successfully route the information from the remaining inputs to infer a full trajectory. The abilities of the approach are demonstrated on synthetic data and real-world data reflecting prototypical object tracking scenarios.
   
%
%
%
\bibliographystyle{splncs04}
\bibliography{Becker_arxiv_2021}

\end{document}